\documentclass[a4paper,twoside]{article}

\usepackage{epsfig}
\usepackage{subcaption}
\usepackage[labelformat=empty]{caption}

\usepackage{calc}
\usepackage{amssymb}
\usepackage{amstext}
\usepackage{amsmath}
\usepackage{amsthm}
\usepackage{multicol}
\usepackage{pslatex}
\usepackage{apalike}
\usepackage[bottom]{footmisc}

\usepackage{multirow}
\usepackage{enumitem}
\usepackage{xcolor, color, soul}
\sethlcolor{yellow}
\usepackage{todonotes}
\usepackage{balance}

\usepackage{xurl}
\usepackage{appendix}

\usepackage[linesnumbered,commentsnumbered,ruled,vlined]{algorithm2e}

\usepackage{SCITEPRESS}     

\renewcommand{\footnotesize}{\fontsize{7pt}{8pt}\selectfont}

\begin{document}

\title{\emph{When the Few Outweigh the Many:\\} Illicit Content Recognition with Few-shot Learning}

\author{\authorname{G. Cascavilla\sup{1}, G. Catolino\sup{2}, M. Conti\sup{3}, D. Mellios\sup{2}, D.A. Tamburri\sup{1}}
\affiliation{\sup{1} Eindhoven University of Technology, Jheronimus Academy of Data Science, The Netherlands}
\affiliation{\sup{2}Tilburg University, Jheronimus Academy of Data Science, The Netherlands}
\email{n(ame).surname@jads.nl}
\affiliation{\sup{3}University of Padova, Italy}
\email{conti@math.unipd.it}
}


\abstract{The anonymity and untraceability benefits of the Dark web account for the exponentially-increased potential of its popularity while creating a suitable womb for many illicit activities, to date. Hence, in collaboration with cybersecurity and law enforcement agencies, research has provided approaches for recognizing and classifying illicit activities with most exploiting textual dark web markets' content recognition; few such approaches use images that originated from dark web content. This paper investigates this alternative technique for recognizing illegal activities from images. In particular, we investigate label-agnostic learning techniques like One-Shot and Few-Shot learning featuring the use Siamese neural networks, a state-of-the-art approach in the field. Our solution manages to handle small-scale datasets with promising accuracy. In particular, Siamese neural networks reach 90.9\% on 20-Shot experiments over a 10-class dataset; this leads us to conclude that such models are a promising and cheaper alternative to the definition of automated law-enforcing machinery over the dark web.}
\onecolumn \maketitle \normalsize \setcounter{footnote}{0} \vfill

\section{Introduction}
\label{intro}

The web as we know it today has two primary layers. On the one hand, the Surface web offers most if not all the web pages we normally use daily. On the other hand, the Deep web---or hidden web~\cite{vldb01raghavan}---offers parts of the World Wide Web whose contents are not indexed by standard web search-engine programs. The latter remains unindexed because its content is considered either irrelevant or confidential, and for security purposes, it is intentionally concealed. The advantages above, however, create a suitable womb for many illicit activities concealed from regular search indexing. Such activities collectively form a relatively small fraction of the Deep web, called the Dark Web~\cite{GodawatteRMS19}. The Dark Web uses the Tor---The Onion Routing\footnote{\url{https://www.torproject.org/}}---network to access its content, and featuring a sensibly different architecture than the Surface web; for example, each request is redirected through various remote servers to reach the requested content and finally return to the user via a different server, thereby making that request untraceable.

Although the Dark Web amounts to circa 0.005\% of the web  \cite{Post_dark_web_analysis}, only 48\% of the Dark Web content is legal \cite{ToRank}, with the rest being illicit, suspicious, or otherwise un-categorized but still within a grey-area of legality (e.g., Smart Drug Trafficking). Such illegal activities usually contain drug selling, counterfeit products, and child abuse content \cite{ChildAbuse}. The majority of these illicit contents are sold through various Dark Web markets. Numerous Surface websites advertise these markets, providing the user with the onion link. Consequently, the Dark Web markets are gaining exponential popularity, endangering, in many cases, the unsuspected user who cannot identify the legality of each product or the truthfulness of the presented information. Lastly, the vulnerability in malicious and phishing code is deep in these markets, posing an additional threat to the everyday user.

In an attempt to shed light on the illicit activities on the Dark Web, the research community is either classifying images, text, or even the underlying code of the dark websites~\cite{cascavilla2022code,secrypt22}. Several studies implement machine learning algorithms and deep learning techniques for automatic taxonomy extraction and Deep and Dark Web content analysis. On the one hand,  image categorization of the HTML pages in the various Dark Web content is researched in depth by \cite{darkvisualpropaganda}. Specifically, authors in \cite{darkvisualpropaganda} identified and categorized dark propaganda based on visual content while using semantic segmentation with specifically designed filters. Finally, \cite{keypointfiltering}, through specially designed masks and ``bags of visual words,'' they classified illicit images from the Dark Web with high accuracy. On the other hand, the textual appearance is the main focus of the studies \cite{al-nabki-etal-2017-classifying} \cite{Categorization_Onion_Sites}. In the latter research \cite{Categorization_Onion_Sites}, they proposed an onion crawler to thematically categorize the content of Dark Web pages, e.g., drug-related, gun-related.

The above studies share one key element, the existence of significantly big datasets that are accurately labeled or, in the case of \cite{keypointfiltering}, a dataset that can be considered ``ideal''. That is adequately cleaned images, lacking any noisy background that someone might encounter when scraping images from the Dark Web. Besides, the data used in the studies mentioned earlier are mostly well-balanced and categorized at a high level, which means that specific categories have not yet been investigated. 
However, ``Reality is cruel,'' meaning relying on more data is not always possible. Law enforcement should have the possibility to react as soon as possible to detect illicit activities on the Dark Web using an approach that works with high accuracy even when the data collection is reduced.
In the context of our research, we provided a novel approach for illicit image recognition, considering new Dark Web images--thus possibly implying small and noisy data. In particular, we investigated an alternative approach when handling small datasets using the ability of label agnostic learning techniques, i.e., One-Shot \cite{One-Shot-Simple-Objects} and Few-Shot \cite{Few-Shot-Agnostic}, when identifying illicit images, thus possibly improving the problem of handling unlabeled and few data. One/Few-shot learning requires fewer data to train a model, thus eliminating high data collection and labeling costs. Moreover, low training data means low dimensionality in the training dataset, which can significantly reduce computational costs. When new data are added, the model can recognize them without re-training. The Dark Web can benefit from these approaches since new illicit content images are arising daily, making their identification time-consuming and challenging. The approaches mentioned rely on using Siamese networks since it can be more robust to class imbalance and works well with images without losing their information. Moreover, it has not been studied on Dark Web content yet.
Consequently, 
we formulated the following main research question:


\begin{center}
    \textbf{RQ.~}\emph{\textbf{To what extent can illicit Dark Web content be classified through a limited number of images?}}
\end{center}

To answer our main research question, we need to address the following sub-questions: 

\begin{enumerate}[start=1,label={SRQ\arabic*.},wide = 0pt, leftmargin = 3em]
\item\textit{To what extent One-Shot technique using Siamese Neural Networks can identify illicit images from the Dark Web?}
\item\textit{To what extent Few-Shot technique using Siamese Neural Networks can identify illicit images from the Dark Web?}
\end{enumerate}
The goal is to investigate the ability of One-Shot and Few-Shot learning techniques to identify and separate illicit image embeddings using Siamese Neural Networks. We verify the results by evaluating the model's performance and focusing mainly on the accuracy metric.

The results of our study highlight that our approach peaked at 90.9 \% testing accuracy on 943 unseen images of 10 different categories.

To sum up, the paper provides four key contributions:

\begin{enumerate}
    \item A novel Dataset of Dark Web illicit contents consisted of 3750 images categorized in 55 different categories, e.g., drugs and weapons~\cite{appendix}.
    
    \smallskip
    \item A new approach that exploits the One-Shot Learning technique to identify illicit images from the Dark Web;
        
    \smallskip
    \item A new approach that exploits the Few-Shot One-Shot Learning technique to identify illicit images from the Dark Web;
    
    \smallskip
    \item An online available repository reporting the raw data in the context of the study for further research and new considerations by the community~\cite{appendix}.
\end{enumerate}

The remainder of this paper is organized as follows. Section~\ref{research_material} provides an overview of the dataset and the related approach used to build, clean, and prepare it. Section~\ref{research_meth} introduces and explains the methodology for classifying illicit images. In Section~\ref{results} are presented the results of our approach. Section~\ref{discussion} discusses the results of our research and the related limitations, while Section~\ref{conclusion} draws the conclusions and sketches some possible future research.

\section{Related Work}
\label{related_w}

Previous research on the Dark Web mainly focused on classifying the illicit activities in the Dark Market places based on their textual content. More specifically, \cite{al-nabki-etal-2017-classifying} created the well-known DUTA dataset, which consists of 5002 labeled Dark websites. Three supervised machine learning algorithms were tested: Support Vector Machines (SVM), Logistic Regression, and Naive Bayes. Using Term Frequency - Inverse Document Frequency (TF-IDF) and Bag Of Words (BOW) dictionaries tuned explicitly for their dataset, they achieved high accuracy when predicting illicit content. Similarly, \cite{Categorization_Onion_Sites} created an onion crawler to thematically categorize the content of Dark Web pages as drug-related, gun-related, etc., based on specific keywords. 
Authors in~\cite{legal_illegal_language}, while following a similar approach, enriched their experiments with data originating from eBay product pages as well as Legal Onion websites in an attempt to identify the legal and illegal language used in the Dark Web. Lastly, \cite{ranade2018using} collected data from the Twitter streaming API to generate a multilingual corpus based on keywords such as DDoS attacks, DNS, spam, malware, etc. The collected data was fed to a translating algorithm designed by the researchers, which achieved 97\% semantic relevance compared to Google's translated output upon expert evaluation.

Even though the textual representations of the Dark Marketplaces are thoroughly investigated, more extensive research should be conducted on the images originating from these markets. One of the most influential studies regarding HTML classification based on the visual contents of Dark websites is \cite{darkvisualpropaganda}. The researchers in \cite{darkvisualpropaganda} are identifying and categorizing dark propaganda based on the visual content of the investigated websites. They trained the well-known Convolutional Neural Network (CNN) Alex-Net on 120,000 images obtained from the Dark Web and finally tested on 1.2 million suspicious images concluding with an accuracy of 86\%. On the other hand, the researchers in \cite{DarkNet-Creic} created a dataset (TOIC) of almost 700 images scraped from the Dark Web. They generated dictionaries representing this database by implementing K-Means and Nearest Neighbour algorithms. Edge-Shifting dense techniques were tested on a different radius, resulting in an 85.6\% overall accuracy. Inspired by the promising results, the authors in~\cite{keypointfiltering} introduced specifically designed masks, and through a similar ``bag of visual words'' BoVW classified illicit images. 
The accuracy of the pre-trained model when tested on the researchers' dataset TOIC while using BoVW reaches approximately 88\%. 

Label-agnostic techniques, such as One-Shot and Few-Shot, learn from the pixels of each image using the Siamese Networks produced embeddings. Therefore, re-training is optional. One of the main differences between One-Shot and Few-Shot Learning techniques is the volume of the input data, which means that the sample of data is used to classify the embeddings produced by the Siamese Networks. In particular, the model is trained on a few images \cite{li2017metasgd} \cite{wang2019generalizing}, or one image per category \cite{shaban2017oneshot} \cite{Vinyals2016MatchingNF}. In~\cite{1467314} are testing the ability of Few-Shot learning implementing Siamese Neural Networks on the AT\&T dataset and the AR database of faces. The datasets, in combination, contain approximately 4000 images of faces photo-shoot in a period of 14 days. Their proposed networks recognize employee faces with an 80\% accuracy.
Siamese Neural Networks is one of the most popular choices for label-agnostic tasks. Its objective is to use twin embedding nets and generate representing vectors for each picture which are compared by calculating their euclidean distance. Studies like in~\cite{Facenet,One-Shot-Object-Categories,One-Shot-Simple-Objects,Omniglot-One-Shot} used the Siamese networks' architecture, obtaining high accuracy in different domains.
All the above studies share one key element, the existence of significantly big datasets that are accurately labeled or, in the case of \cite{keypointfiltering}, a dataset that can be considered ``ideal'', where images are cleaned and lacked from noisy. Also, the data used are mostly well-balanced and categorized at a high level, which means that specific ``in-depth'' categories have not been investigated yet. Therefore, we investigated an alternative approach when handling small datasets using the ability of label agnostic learning techniques, i.e., One-Shot \cite{One-Shot-Simple-Objects} and Few-Shot \cite{Few-Shot-Agnostic}, when identifying illicit images, thus possibly improving the problem of handling unlabeled and few data. To the best of our knowledge, no previous research investigated the ability of label-agnostic techniques for illicit image recognition, which is the focus of our work.
\section{Dataset Overview and Data Engineering}
\label{research_material}

This section reports the steps followed to extract new images from the Dark Web and create our datasets~\cite{appendix}.

\subsection{Data Scraping - Collection}
\label{sec}
To collect data, we implemented a crawler using the Selenium Python library capable of automating steps to download HTML pages from the Dark Web using the Tor browser. 

Since login into the website was mandatory to extract any information, we needed to deal with the security Captchas using Captcha-solving API. The script captured a screenshot of the website's login page, which was sent to the external server. After locating the input box, the resulting password was automatically typed into the appropriate field. 
 
After logged in, the script crawled through the different product ads and collected the URLs of the images. Initially, the objective was to download the images of the products immediately after redirecting to the product page. However, this technique was identified as an attack and blocked. Hence, we built a list with external links of all the product images accompanied by the category these images belonged to. Lastly, a different script bypassed, in a similar manner, the security of the website and randomly downloaded the images from the servers, avoiding triggering any alarms. The data used in this research have been scraped in a period between January and March 2020 from various Dark Markets. More specifically, 
we scraped three popular Dark Web marketplaces Silk Road, BitBazaar, and Dark Market, resulting in 5500 images depicting drugs of all categories, credit cards, ID cards (IDs), and gift cards. Although the markets above broadly related to drugs, the sample of personal IDs and credit cards was relatively small, while the sample of passports was less than five images. These Dark markets also lacked images of weaponry, so we scraped additional random onion sites resulting in 210 high-quality images of guns and semi-automatic guns, 215 ID cards, 51 images of passports, and 118 additional credit cards. 
The dataset with all the data is currently stored in an encrypted hard drive and available under request. However, it is worth highlighting that the authors did not buy any item advertised in the markets cited above. All the images are publicly available from the crawled dark marketplaces as product advertisements.

 
\subsection{Data Cleaning and Preparation}
\label{data_preparation}
To prepare our data, we performed common steps like cleaning and removing duplicates.
In particular, we tested for identical duplicates through hashing techniques\footnote{Image Hashing - Python Documentation: \url{https://pypi.org/project/ImageHash/}}, resulting in about 2000 matching images. For this reason, we removed them from the dataset. In the online appendix \cite{figure_appendix} Figure~\ref{fig:Dataset_Distribution} shows the distribution of the dataset.

Finally, based on the availability problem described in \ref{sec}, we dealt with merging, removing, or relabeling specific sub-categories (the removed categories are marked with a red dot in Figure \ref{fig:Dataset_Distribution} (in online appendix~\cite{figure_appendix}),  Moreover, we created a new category of counterfeit including passports, IDs, money bills, credit cards, gift cards, and documents. All the drug-related categories kept their initial labels. All these steps concluded in a dataset of 3570 images and 55 different classes.

\subsection{Data Augmentation}
\label{data_augmentation}

\begin{figure}[h!]
\centering
 \includegraphics[width=0.47\textwidth]{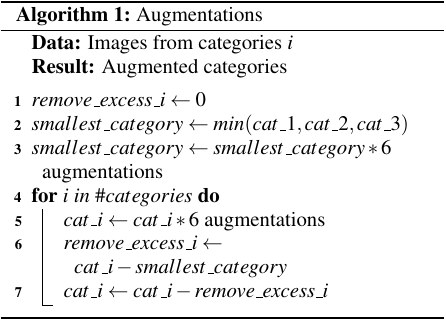}
 \caption{}
 \label{fig:algorithm2}
 \end{figure}

Even after the various cleaning steps and precise categorization, the final dataset results unbalance. In the context of our paper, we experiment with our approach using One-Shot \cite{One-Shot-Object-Categories} and Few-Shot \cite{wang2019generalizing}. Since previous studies (\cite{o2019one,ochal2021class}) advise using them on balance data--with an identical sample of images for each category--avoiding a poor representation of specific categories, we performed data augmentation to balance the minority categories, e.g., type of drugs. In particular,  the script calculates the final size after the possible augmentation steps and aids the user in a better sample decision. The code augments each image six times and saves it for later use while randomly removing excess images from the more significant categories to balance them with the remaining ones. 
The volume of images that needed to be removed was calculated based on the size of the smallest size category, and the images were deleted from each category. 
Pseudocode is available in Algorithm~\ref{fig:algorithm2}.


\begin{figure}[h!]
\centering
 \includegraphics[width=0.47\textwidth]{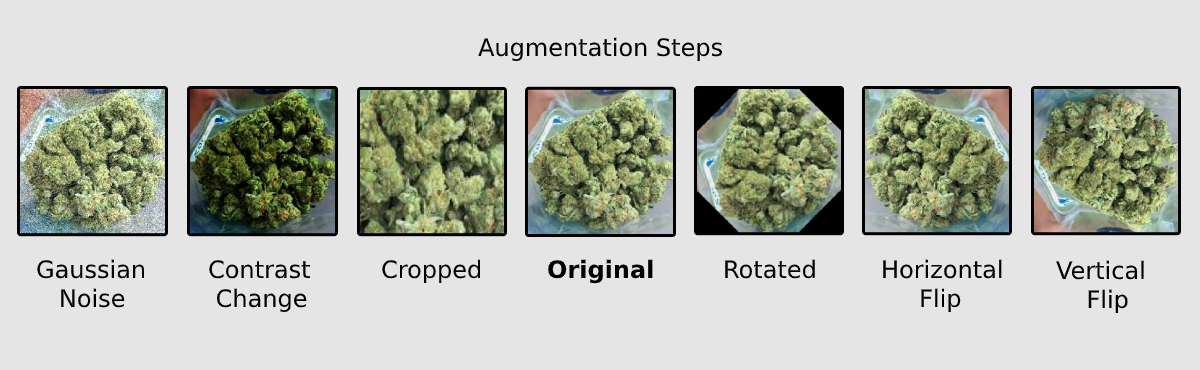}
 \caption{The 6 different augmentations implemented
to this image from left to right are: Gaussian Noise of density 30, Contrast Change (gamma = 2.4), Zoom-Crop by 30\%, Rotation by 40 degrees, Horizontal flip, and Vertical Flip.}
 \label{fig:Augmentations}
 \end{figure}

The augmentations steps are: rotation by 30 degrees, horizontal flip, vertical flip, cropping by 30\%-45\%, change of contrast's gamma by 2.0 - 3.0, and addition of Gaussian noise. The augmentations are illustrated in Fig.~\ref{fig:Augmentations}. We generated the augmentations using Imgaug Augmenters\footnote{Documentation and Examples:~\url{https://imgaug.readthedocs.io/en/latest/source/overview/arithmetic.html}}. 

\section{Research Methodology}
\label{research_meth}
This section explains the methodology implied to classify illicit images and carry out our experimental evaluation.

\subsection{Experimental setup}      
\begin{figure*}[h!]
\centering
 \includegraphics[scale=0.28]{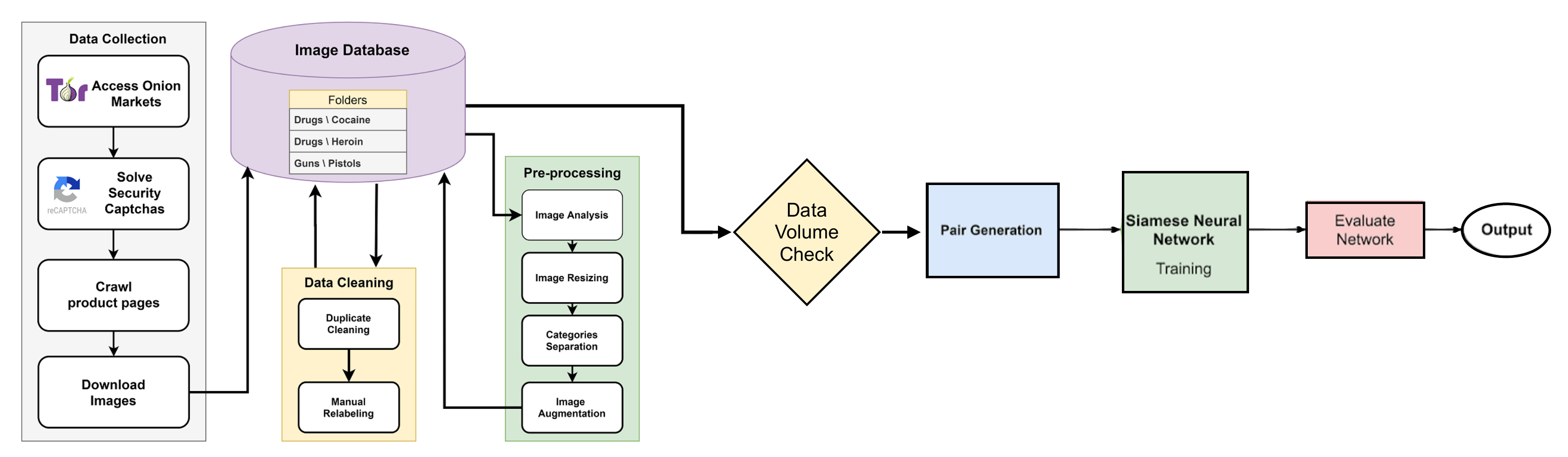}
 \caption{Pipeline of the study.}
 \label{fig:Overall_pipeline}
 \end{figure*}

Figure \ref{fig:Overall_pipeline} reports the experimental research pipeline followed in our study.
The first step regards using the scraping tool that accesses the Dark Web through the Tor browser to scrape Onion websites. The images were randomly downloaded and stored locally in the appropriate folders based on their category, we removed duplicates, and some images were relabeled manually for better representation. We evaluated the most popular dimensions of images in the dataset and informed the user appropriately. We augmented the images and balanced the classes based on the users' dictation. Finally, we trained and evaluated the Siamese neural network using One-Shot and Few-Shot learning.

We implemented our script and model using Python using libraries like Selenium \cite{GOJARE2015341}, Glob\footnote{Glob Documentation: \url{https://docs.python.org/3/library/glob.html}}, Shutil\footnote{Shutil Documentation: \url{https://docs.python.org/3/library/shutil.html}}, Image PIL\footnote{Image PIL: \url{https://pillow.readthedocs.io/en/stable/reference/Image.html}}, ImGaug Augmenters\footnote{Augmentation Library: \url{https://imgaug.readthedocs.io/en/latest/}}, Tensorflow\footnote{Tensorflow: \url{https://www.tensorflow.org}}, Sklearn\footnote{Sklearn: \url{https://scikit-learn.org/stable/}}, and TSNE \cite{lin2017transfer}.

\subsection{One-Shot and Few-Shot learning}
The One-Shot and Few-Shot techniques merely differ in the volume of the input data used for creating the embeddings and testing the models. In other words, one or a few images are used for each category for the One-Shot and the Few-Shot experiments. The number of categories is represented by k, hence, k-way datasets and N-shot where N is the number of images in each category.

\subsection{Pair Generation for Siamese Networks}
 
Before we move on to the models, it is essential to explain the pair generation procedure in detail. K-pairs of N images must be generated to test the Siamese Neural Networks on K-Shot experiments. Hence the higher the sample of data in each category, the more pairs can be generated. In particular, each image is randomly paired to 1,2,5(etc.) images, as described in \cite{varior2016gated} \cite{Omniglot-One-Shot} \cite{qiao2017fewshot}, and a binary label is assigned to the pair. If both images of the generated pair belong to the same category, the label is 1; otherwise is 0. However, randomly created pairs generally produce an imbalanced representation of the positive (1) and negative (0) pairs. For example, if there are 10 images in 10 different categories, and we choose one image from the first category with the goal being to find another image from the same category, the probability of achieving that is \(\frac{9}{99}\) or 0.0909. That number is only decreasing \(\frac{9}{99}*\frac{8}{98}*\frac{7}{97}*.. \) when we try to randomly select more images from the same category for a 5-Shot approach. Therefore, the positive pairs are disproportionately less than the negative.  

In this paper, the generated pairs maintain the same number of negative and positive images similarly designed to \cite{shaban2017oneshot}, even though the pairs were generated randomly. For each experiment, we first created the positive pairs, followed by the same number of negative ones, reassuring an accurate representation of the two labels [1,0] and eliminating any label bias. Based on this study's testing, the final accuracy can fluctuate drastically if the preparation of the pairs is not designed correctly. Meaning that the model will search for the easiest solution to produce the highest accuracy. That is, the output embeddings are always far away from the compared ones, and the model tends to predict a label 0 on every set because it cannot penalize the mistakes adequately. A solution to this issue follows the pair creation of \cite{varior2016gated}. Instead of searching for an image from the same category, the authors generated the positive pair by augmenting the initially chosen picture. Nonetheless, in this paper, the chosen images are always different from the comparing one because the augmentation is applied manually and in an earlier stage, as described in Section~\ref{data_augmentation}. This way, the compared embedding is rarely very close, constituting even harder One-Shot and Few-Shot tasks.

\subsection{Siamese Neural Network Architecture}
We created and tested different Siamese Neural Networks to identify the optimal number of hidden layers needed and the activation functions for the hidden and output layers. Also, we varied the last fully connected (Dense) layer and the in-between filter sizes to create a model able to extract the embedding of each input image pair as accurately as possible. The difficulty of the task highly correlates with the depth of the neural network. Our proposed model does not strictly follow any of the networks proposed in ~\cite{One-Shot-Simple-Objects,Few-Shot-Agnostic,Omniglot-One-Shot}, but it is highly inspired by several related studies \cite{qiao2017fewshot,li2017metasgd,Vinyals2016MatchingNF}. We tuned the final architecture of the convolutional embedding neural networks on the datasets created. The proposed structure of the convolutional embedding neural network consists of six convolutional 2D layers. The first two layers have a filter size of 3 by 3, and the remaining of 2 by 2. The layers dimensions are increasing gradually, starting from 50 by 50 up to 220 by 220. 

Between the convolutional 2D layers, we applied a max-pooling of (3x3) and (2x2) as depicted in \cite{varior2016gated}. The filters of the hidden convolutional layers were tested with various sizes to eliminate overfitting effects. Additionally, we applied three lasso regularizations to the fully connected layer (512); a kernel regularization of 0.001, a bias regularization, and an activity regularization. The last Dense layer of the Embedding Network has a ReLu activation function which, compared to a linear one, produced more stable results. Lastly, the model implements the RMSprop optimizer with a 0.0001 learning rate and a decay of 0.7.
The final structure of the Siamese network consists of two identical embedding neural networks and is illustrated in Fig.~\ref{fig:Embedding_Net} (in online appendix~\cite{figure_appendix}).


Each embedding network is fed with one of the generated pairs' images. The twin embedding networks (Fig.~\ref{fig:Siamese-Structure} in the online appendix~\cite{figure_appendix}) are fully connected to the final dense layer that outputs either 1 if the pair is originated from the same category or 0 if it is from a different one. The Siamese Neural network predicts the label by calculating the Euclidean Distance between the two embeddings, as depicted in \cite{7899663} and \cite{varior2016gated}. Lastly, the network calculates the loss per pair of images via the Contrastive loss formula, as depicted in \cite{hadsell2006dimensionality}. 
In this research, similarly to previous studies, K-shot tests were performed. Therefore, we generated 1 pair and 5 pairs per image. The Siamese neural network is tested on 1-Shot and 5-Shot learning approaches with various samples of illicit pictures for each category. Besides, the networks were tested with more and fewer categories for each K-shot technique.

\section{Results}
\label{results}

We performed two types of tests to evaluate the ability of the Siamese neural network regarding proper embedding creation and accurate separation of them.
The models are tested on three different buckets of data. In particular, we experimented with our models considering the shape and the actual type of illicit content images. Following the experimentation techniques of \cite{garcia2017fewshot} and \cite{Vinyals2016MatchingNF}, the tests are performed on gradually increased datasets. In particular, the first bucket consists of 55 categories (55-way) of illicit images; each category is represented by just one image. The second and third buckets use the same dataset, but each category includes 5 and 20 images, respectively.
Additionally, the number of classes in the buckets above was randomly reduced to identify the model's ability to separate a higher variety of embeddings. Hence, the model is also trained on 10, and 25 randomly selected categories. The tests were performed with 943 randomly chosen entirely new images, and the models were trained for 100 epochs.

Table \ref{table:Acc_Catalogue} illustrates the accuracy of the various tests performed. Specifically, each model is tested on three category volumes, 10-way, 25-way and 55-way and for 1-Shot, 5-Shot, and 20-Shot tests. Looking at the table, the model was fed with 1 image per category, then 5 images per category, and finally 20 images. As expected, the model performs better when tested with a higher N-Shot since there are more trainable examples per category. Generally, 1-Shot tests were prone to overfitting, whereas the overfitting effect was drastically reduced in the 5-Shot and 20-Shot tests.

We can claim that the Siamese Neural Network results in higher training and validation accuracy when more data are present. In addition, the testing accuracy depicts an increase of 20\% (from 70.1\% to 90.9\%) if 20 images are present (20-shot) in 10 classes (10-way), compared to just one image in each one of them. The same pattern is visible throughout the different category sizes, with approximately 19\% (from 66.8\% to 86.7\%) increase in the 25 classes (25-way) test and 14.8\% (from 71.4\% to 86.2\%) in the 55 classes test. These results imply that the model can generalize better due to the increased size of the trainable examples. Additionally, the model performs better when the categories are reduced from 55 to 25 and 25 to 10, but the difference never exceeds 4.2\%. Meaning that the model is not affected by the number of classes if the volume of images in each class does not exceed the above sizes. It is worth noticing that in the case of 1-shot, the model under-performs on testing when the number of classes is reduced from 55 to 25. That occurs because the classes are randomly split and reduced; therefore, some classes might be recognized with higher precision in the 55-way bucket. 

\begin{figure*}[h]
\centering
\subfloat[ROC curve: 55-way on 1-shot ]{\label{fig:Roc_curve1}{\includegraphics[width=0.33\textwidth]{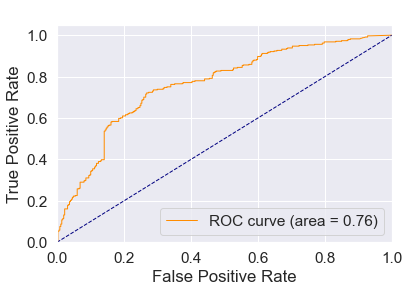}}}\hfill
\subfloat[ROC curve: 55-way with 5-shot ]{\label{fig:Roc_curve2}{\includegraphics[width=0.33\textwidth]{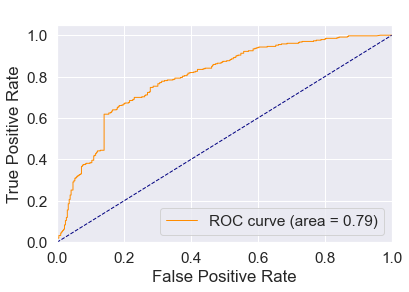}}}\hfill
\subfloat[ROC curve: 55-way on 20-shot ]{\label{fig:Roc_curve3}{\includegraphics[width=0.33\textwidth]{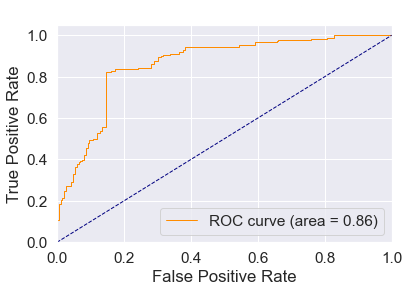}}}
\caption{The ROC curves of the 5-Shot learning tests on 55-way with 1-shot, 5-shot, and up to 20-shot respectively.}
\label{fig:Roc_curves}
\end{figure*}

The ROC curves of the N-Shot tests conducted on the 55-way bucket of data in Fig.~\ref{fig:Roc_curves} are in charge of further justifying the increase in generalization when more data are present in each category. Looking at sub-figures \ref{fig:Roc_curve1} and \ref{fig:Roc_curve2}, it is safe to conclude that when additional images are present in each of the classes, while training, the Siamese Network can identify more tested positive pairs/labels. The above is visible in the ROC curve area of the last sub-figure \ref{fig:Roc_curve3}, which is equal to 86\%, approximately 10\% higher compared to the first \ref{fig:Roc_curve1}.

\begin{table}[h]
\centering
\resizebox{0.9\linewidth}{!}{
\begin{tabular}{lllll}
                &               &                 &                 &                  \\
                &               & \textbf{1-Shot} & \textbf{5-Shot} & \textbf{20-Shot} \\
                &               &                 &                 &                  \\ \hline
\textbf{10-way} & Val Accuracy  & 98.9\%         & 93.8\%         & 96.4\%          \\
                & Test Accuracy & 70.1\%         & 75.6\%         & \colorbox{lightgray}{\textbf{90.9}\%}          \\
                &               &                 &                 &                  \\ \hline
\textbf{25-way} & Val Accuracy  & 98.7\%         & 92.7\%         & 92.1\%          \\
                & Test Accuracy & 66.8\%         & 76.2\%         & \colorbox{lightgray}{\textbf{86.7}\%}          \\
                &               &                 &                 &                  \\ \hline
\textbf{55-way} & Val Accuracy  & 97.6\%         & 86.2\%         & 87.7\%          \\
                & Test Accuracy & 71.4\%         & 74.3\%         & \colorbox{lightgray}{\textbf{86.2}\%}

\end{tabular}
}
\caption{The accuracy of the various tests with 10-way, 25-way and 55-way dataset on 1-shot, 5-shot, and up to 20-shot respectively.}
\label{table:Acc_Catalogue}
\end{table}

\section{Discussion}
 \label{discussion}
 This section discusses the results and limitations of the study.

\subsection{Research Questions}
The main objective of this research was to investigate alternative approaches that can recognize illicit activities on the Dark Web. Specifically, this study aimed to bypass the burden of collecting supervised large-scale datasets using One-Shot and Few Shot learning. 
Therefore, the main question was related to the ability to detect illicit Dark Web content with a limited number of images. Our experiment showed promising results. Indeed, we can claim that Siamese Neural Networks can recognize illicit images efficiently based on this study's experimentation when considering the first subquestions related to using One-Shot learning techniques. Indeed when testing the accuracy of the model, we reach a percentage around 70\%. 
Moving the attention to Few-Shot learning techniques, i.e., SRQ2, the Siamese Neural Network presented promising generalization capabilities when the sample was increased by just four or up to 20 images per category. The testing accuracy reached 76.2\% on the 5 images per category dataset with 25 categories and 90.9\% on the 6 to 20 images per category dataset with ten categories. Lastly, it is worth noticing that the testing accuracy stayed under 86.2\% regardless of the increased number of categories, 55, on the 6 to 20 images dataset.
These techniques' usage can be promising compared to the previous study. In particular,  Fidalgo et al. \cite{DarkNet-Creic} proposed an approach for detecting illicit contents on a small-scale dataset of approximately 700 images separated in 5 classes. The resulting accuracy in \cite{DarkNet-Creic} is 85.6\% and in \cite{keypointfiltering} 87.98\%. Our study outperforms the aforementioned by 3\% with a 90.9\% on 20-Shot tests. Finally, we scored similar results, 86.7\%, with a dataset consisting of 25 classes, 20 classes more than the previous research. Our future agenda aims at comparing the methodology of their study.

\subsection{Limitations}
To the best of our knowledge, no previous study has investigated the above techniques on illicit content. Hence, a strait forward comparison of this study's results with previous studies was impossible.
Finally, precise categorization and data cleaning were among the initial burdens. Random pair generation was the only possible solution, even though previous studies suggest a manual selection of them. Based on our study's results, higher precision of labeling yields superior accuracy. Although, due to the lack of expertise regarding illicit content labeling, the categories could not be further separated, and the pairs could not be generated manually. To avoid possible problems in the code, we developed our script and pipeline, we relied on stable Python Libraries.

\section{Societal Impact}

This study aims to investigate alternative approaches when handling small datasets. The expensive time procedure of collecting large-scale data (images) from Dark Web Markets, as well as the need for highly skilled personnel responsible for illicit content labeling, are some of the burdens this research is trying to bypass. We showed the ability of label-agnostic models handling unlabeled data to identify illicit images from the Dark Web. Law enforcement agencies can benefit from our suggested approach and could conduct faster investigations with fewer resources and capabilities. Moreover, the recognition speed of new illegal substances from the Dark Web represents a key factor in intercepting new illegal trends and persecuting illicit behaviors. Our proposed approach poses the basis for a less time-consuming system to assist law enforcement agencies during their activities.

\section{Conclusion}
\label{conclusion}

This study presents a novel approach to recognizing illicit images from the Dark Web through a relatively small sample of images. We generated a new dataset consisting of 3570 images spreading over 55 sub-classes.
Then, we investigated the Siamese neural network classification methods on One-Shot and Few-Shot experiments. Results show that Siamese network peaked at 90.9\% testing accuracy on 943 unseen images of 10 different categories.
To conclude, this study provided a new contribution to illicit image recognition through label-agnostic networks on One-Shot and Few-Shot experiments. These techniques could help law enforcement agencies effortlessly identify illicit activities in the Dark Web through small data samples. The future agenda includes the comparison of other different techniques and setting parameters.

\balance
\bibliographystyle{apalike}
\small
\bibliography{biblio}

\newpage
\appendix

\begin{figure*}[h!]
\centering
 \includegraphics[width=0.9\textwidth]{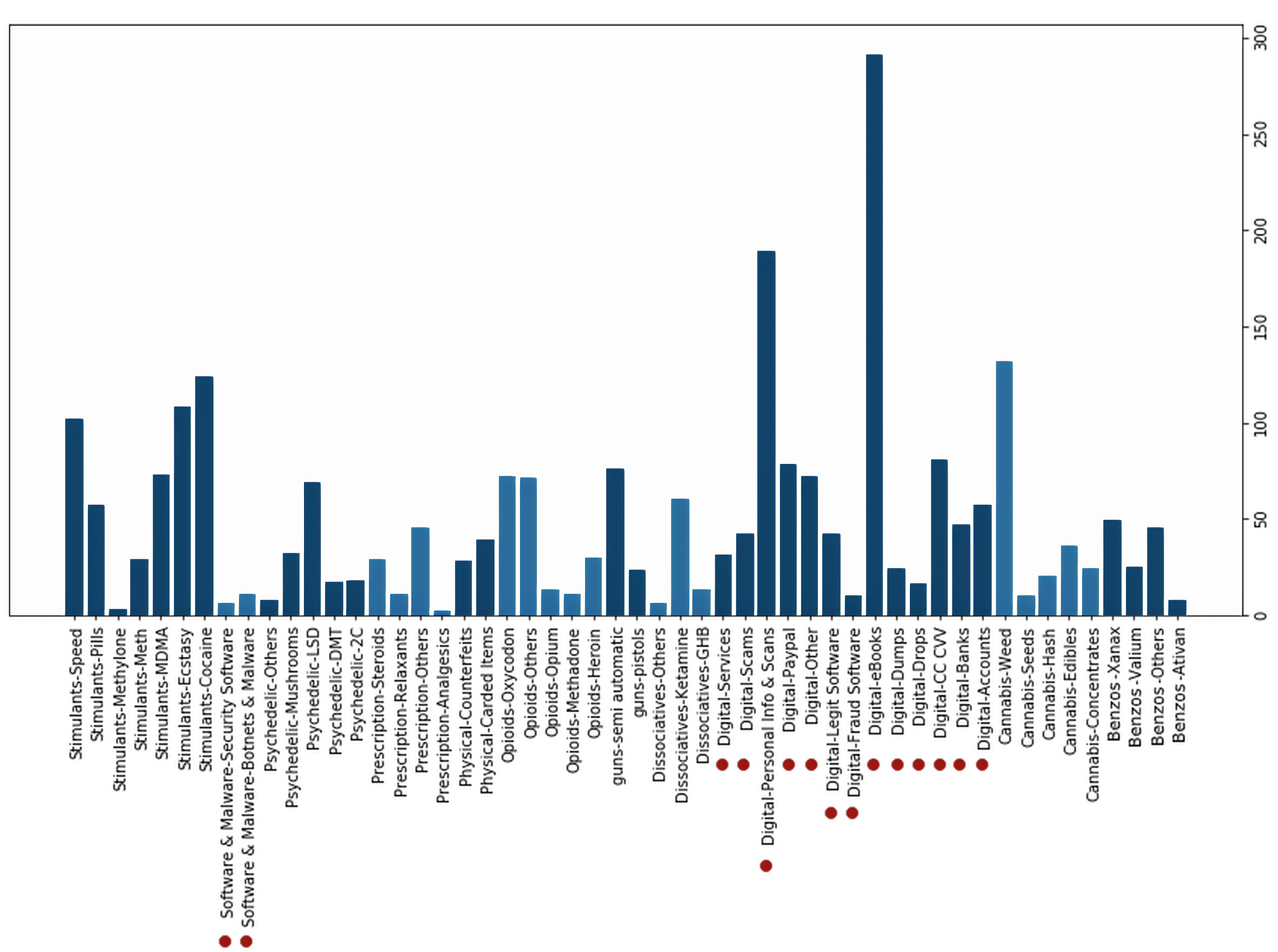}
 \caption{The distribution of the initial dataset. The red dots represent the categories that are excluded from the final experiments.}
 \label{fig:Dataset_Distribution}
 \end{figure*}

 \begin{figure*}[h]
\centering
 \includegraphics[scale=0.15]{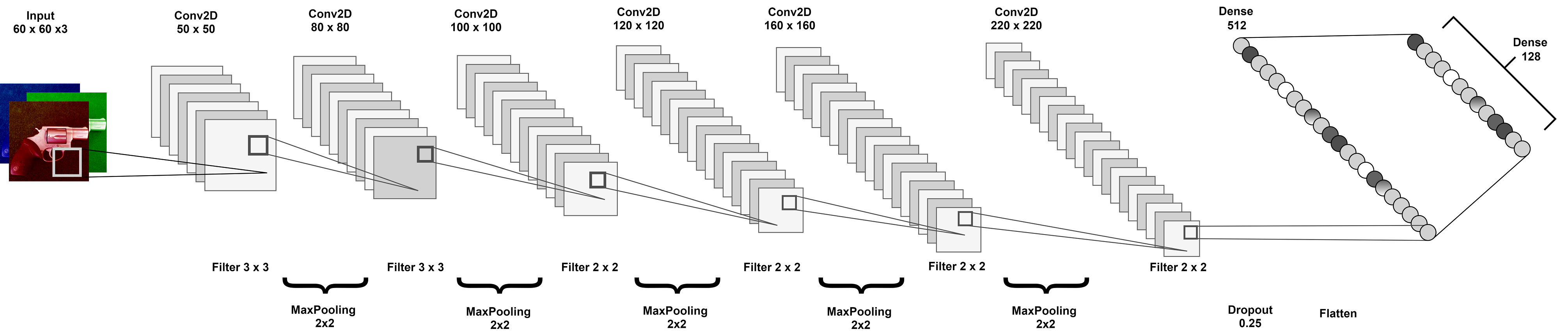}
 \caption{The proposed embedding convolutional neural network. A Siamese network consists of two identical embedding nets.}
 \label{fig:Embedding_Net}
 \end{figure*}

  \begin{figure*}[h]
\centering
 \includegraphics[width=\textwidth,height=0.4\textwidth]{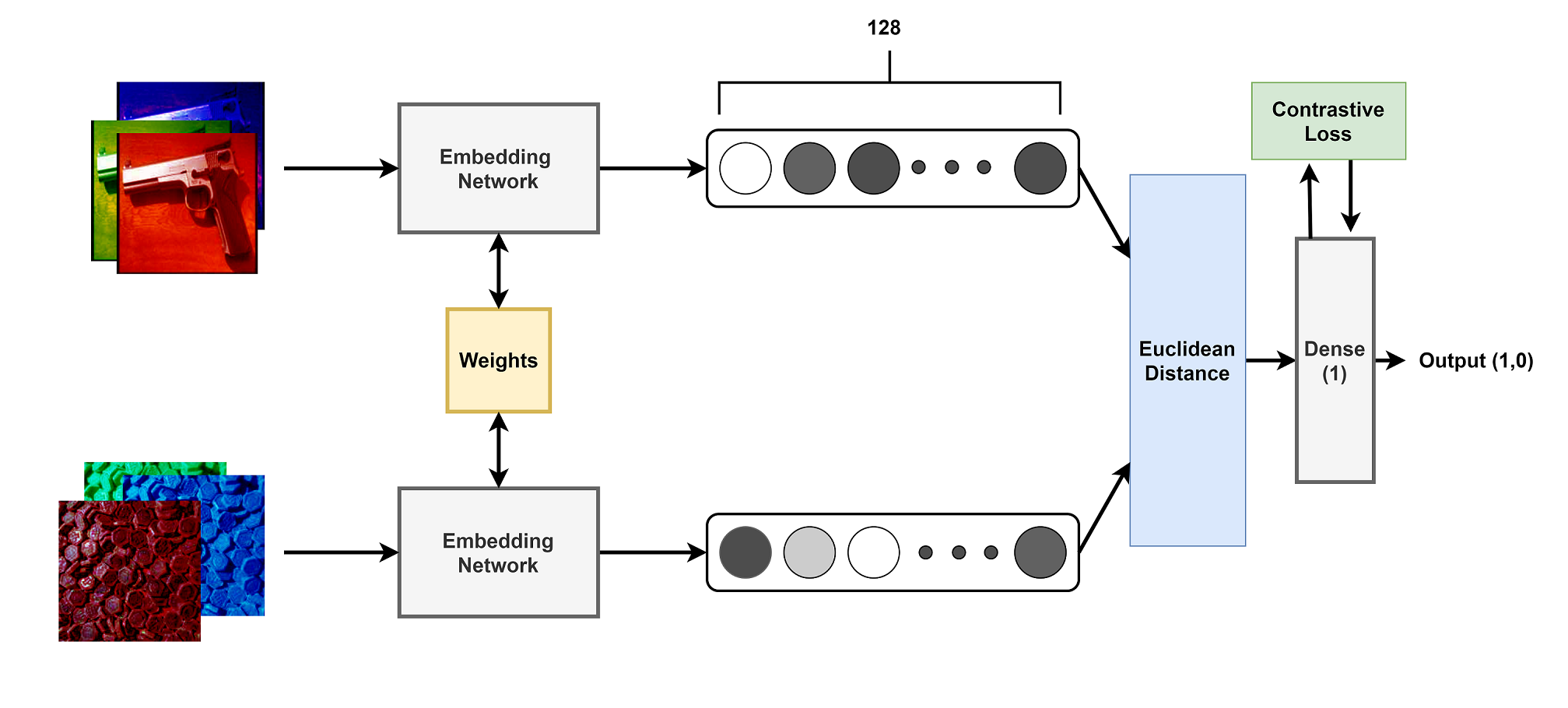}
 \caption{The twin embedding networks (Siamese Neural Network). The weights are shareable between the twin networks at the last fully connected layer. The output size for each embedding is 128. A fully connected layer is outputting 1 or 0 based on the error calculated from the Contrastive loss.}
 \label{fig:Siamese-Structure}
\end{figure*}

\end{document}